\pdfoutput=1
\documentclass[11pt]{article}

\usepackage[final]{acl}

\usepackage{times}
\usepackage{latexsym}
\usepackage{amsmath}
\usepackage{amsfonts}
\usepackage{amssymb}
\usepackage{mathtools}
\usepackage{bm}
\usepackage{microtype}
\usepackage{inconsolata}
\usepackage{siunitx}
\usepackage[T1]{fontenc}
\usepackage[utf8]{inputenc}
\usepackage{graphicx}
\usepackage{booktabs}
\usepackage{longtable}
\usepackage{array}
\usepackage{geometry}
\usepackage{multirow}
\usepackage{caption}
\usepackage{enumitem}
\usepackage{algorithm}
\usepackage{algorithmic}
\usepackage{xspace}
\usepackage{url}

\usepackage{float} % for [H] if allowed
\usepackage{caption}
\newcommand{\indicator}[1]{\mathbf{1}\!\left[#1\right]}

\newcommand{\method}{SD-E$^{2}$\xspace}   % our method
\newcommand{\grpocfee}{GRPO-CFEE\xspace}                      % baseline

\newcommand{\grpoCFL}{GRPO-CFL\xspace}

\newcommand{\gsmBase}{54.66}

\newcommand{\gsmSemDivAcc}{\textbf{82.03}}     % TODO
\newcommand{\STRAT}{\langle \mathrm{STRAT}\rangle}
\newcommand{\FA}{\langle \mathrm{FA}\rangle}      % final_answer
\newcommand{\ANS}{\langle \mathrm{ANS}\rangle}    % answer
\newcommand{\SO}{\langle \mathrm{SO}\rangle}      % strategy_outcome
\newcommand{\fans}{f_{\mathrm{ans}}}
\newcommand{\clip}{\mathrm{clip}}
\newcommand{\Uniq}{\mathrm{Uniq}}
\newcommand{\Divv}{\mathrm{Div}}
\newcommand{\Div}{\mathrm{Div}}              % diversity

\newcommand{\KL}{D_{\mathrm{KL}}}

\title{\method: Semantic Exploration for Reasoning Under Token Budgets}

\author{
  Kshitij Mishra \qquad Nils Lukas \qquad Salem Lahlou \\
  Mohamed bin Zayed University of Artificial Intelligence \\
  \texttt{\{kshitij.mishra,nils.lukas,salem.lahlou\}@mbzuai.ac.ae}
}

\begin{document}
\maketitle

\begin{abstract}
Small language models (SLMs) struggle with complex reasoning because exploration is expensive under tight compute budgets. We introduce Semantic Diversity – Exploration–Exploitation (\textbf{\method}), a reinforcement learning framework that makes exploration explicit by optimizing \emph{semantic} diversity in generated reasoning trajectories. Using a frozen sentence-embedding model, \method assigns a diversity reward that captures (i) the coverage of semantically distinct solution strategies and (ii) their average pairwise dissimilarity in embedding space, rather than surface-form novelty. This diversity reward is combined with outcome correctness and solution efficiency in a $z$-score–normalized multi-objective objective that stabilizes training. On GSM8K, \method surpasses the base \texttt{Qwen2.5-3B-Instruct} and strong GRPO baselines (\grpoCFL\ and \grpocfee) by ~+27.4, ~+5.2, and +1.5 percentage points, respectively, while discovering on average 9.8 semantically distinct strategies per question. We further improve MedMCQA to 49.64\% vs 38.37 for base and show gains on the harder AIME benchmark (1983--2025), reaching 13.28\% vs. base 6.74\%. These results indicate that rewarding semantic novelty yields a more compute-efficient exploration–exploitation signal for training reasoning-capable SLMs. By introducing cognitive adaptation (adjusting the reasoning process structure rather than per-token computation), \method offers a complementary path to efficiency gains in resource-constrained models.

% On GSM8K \method outperforms base model -- \texttt{Qwen2.5-3B-Instruct},  strong baselines \textit{viz.} \textbf{\grpoCFL} (GRPO with the reward design of \cite{deepseek2025r1}) and \textbf{\grpocfee} (GRPO with semantically agnostic exploration-exploration rewards), with accuracy gains of 26\%, 6\% and 1.5\% respectively, while uncovering a set of on an average 9.8 reasoning strategies. These results indicate that rewarding semantic novelty provides a more compute-efficient exploration–exploitation signal for training reasoning-capable SLMs.
\end{abstract}

\section{Introduction}

Large Language Models (LLMs) have demonstrated remarkable reasoning ability across mathematics, science, and general-domain tasks~\citep{wei2022chain,kojima2022large,bubeck2023sparks,yao2023tree,zelikman2023parsel}. 
Techniques such as Chain-of-Thought prompting~\citep{wei2022chain} and Tree-of-Thoughts search~\citep{yao2023tree} enable these models to generate multi-step reasoning traces and explore alternative strategies. 
However, their immense scale---often tens or hundreds of billions of parameters---comes with high inference cost and latency, motivating a shift toward \textbf{Small Language Models (SLMs)} for cost-efficient and deployable reasoning~\citep{chen2023phi2,team2024phi3}. 
Yet, SLMs struggle to match the reasoning fidelity of their larger counterparts. 
Their limited capacity increases susceptibility to exposure bias~\citep{ranzato2015sequence}, while their tight token budgets constrain the complexity and length of reasoning paths.

\begin{figure*}[t]
  \centering
  \captionsetup{font=footnotesize} % <-- makes this one caption smaller
  \includegraphics[width=\textwidth,keepaspectratio]{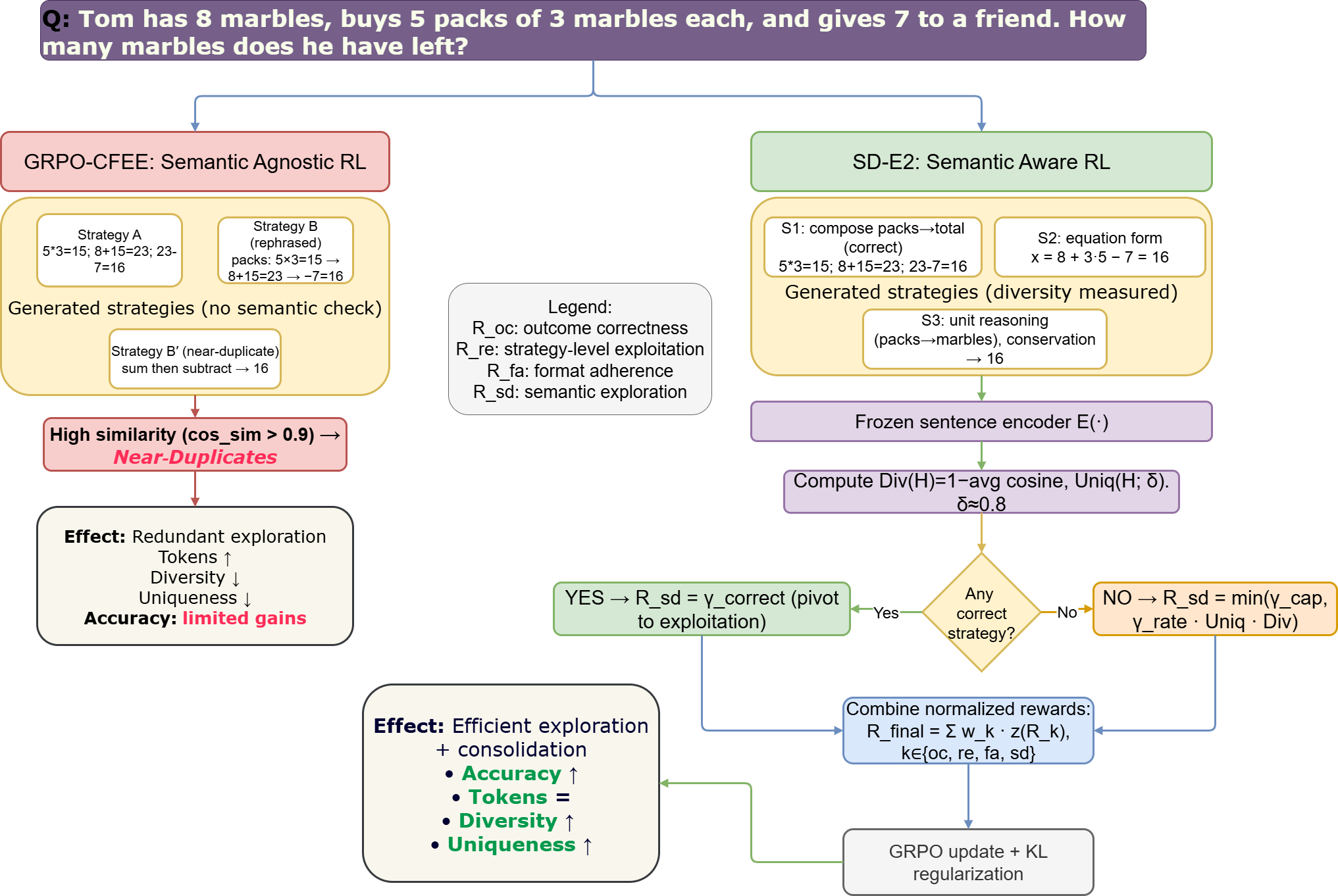}
  \caption{\textbf{Problem and approach overview on a GSM8K example.}
  \emph{Left:} Outcome-driven baselines (e.g., \grpoCFL) and the non-semantic \grpocfee can generate
  multiple, near-duplicate strategies, leading to redundant exploration (Tokens~↑, Diversity~↓).
  \emph{Right:} \method encodes each \texttt{<reasoning>} with a frozen sentence encoder to compute
  (i) $\mathrm{Div}(H)=1-\text{avg cosine}$ and (ii) $\mathrm{Uniq}(H;\delta)$, rewarding exploration
  only when strategies are \emph{semantically} distinct; upon any correct strategy, it switches to an
  exploitation bonus. Normalized components $R_{\mathrm{oc}},R_{\mathrm{re}},R_{\mathrm{fa}},R_{\mathrm{sd}}$
  are combined under a GRPO objective with KL regularization, yielding higher ACC under the same max-token decoding budget.}
  \label{fig:semdiv-intro}
  \vspace{-0.6em}
\end{figure*}

This limitation introduces a fundamental tension between \emph{exploration} and \emph{exploitation}. 
An SLM must explore diverse reasoning strategies to escape local optima and discover valid solution paths, yet it must quickly exploit promising avenues to stay within its computational and token budget. 
Existing methods inadequately resolve this trade-off. 
Inference-time ensembling techniques such as Self-Consistency~\citep{wang2022selfconsistency}, Tree-of-Thoughts~\citep{yao2023tree}, and Reasoning-as-Planning~\citep{zhou2023lats} improve accuracy but incur significant overhead, negating the efficiency gains of smaller models. 
Meanwhile, Reinforcement Learning (RL) alignment methods such as RLHF~\citep{christiano2017deep,bai2022trainingHF,ouyang2022training} and preference-optimization variants like DPO~\citep{rafailov2023direct}, IPO~\citep{garg2025ipo}, and GRPO~\citep{shao2024deepseekmath} rely primarily on sparse outcome-based signals (e.g., correctness or preference). 
Recent works on process supervision~\citep{lightman2023let,wang2023math0shepherd0,huang2024frost,zhou2025sweet} take a step further by rewarding intermediate reasoning steps, yet they still lack a measure of \emph{exploration quality}. 
As a result, current methods cannot distinguish between discovering a genuinely novel reasoning strategy and merely rephrasing an existing one, leading to repetitive and inefficient search behavior.

In this work, we introduce \textbf{\method}, a semantics-aware reinforcement learning framework that teaches SLMs to reason efficiently by rewarding exploration only when it is \emph{meaningfully different}. 
At its core is a \textbf{semantic exploration reward} that leverages a frozen sentence-embedding model to measure the diversity of reasoning traces. 
When a correct solution is found, \method\ shifts focus to exploitation through a fixed reward bonus, encouraging consolidation of success. 
When no correct strategy is discovered, the exploration reward scales with both the number of \emph{semantically unique} reasoning paths and their average dissimilarity, promoting broad yet targeted exploration of novel ideas rather than superficial rewording.
This represents a form of \emph{cognitive adaptation}~\citep{graves2016adaptive}: rather than adapting the per-token computational cost through architectural means (e.g., early exiting, sparse experts), \method adapts the high-level reasoning process itself based on semantic saturation, preventing the generation of entire redundant strategy blocks.

As summarized in Fig.~\ref{fig:semdiv-intro}, baselines without a semantic signal often produce
near-duplicate strategies (high cosine similarity), while \method measures semantic diversity
and rewards only \emph{meaningfully different} exploration, pivoting to exploitation once any
strategy yields the correct outcome.

On GSM8K, we first compare against the base model: with identical prompts on \texttt{Qwen2.5-3B-Instruct}, \method\ improves accuracy by \textbf{+27.4} points. We then benchmark against strong GRPO baselines (outcome-driven GRPO~\citep[\grpoCFL;][]{deepseek2025r1}) and a non-semantic explore--exploit baseline (\grpocfee), obtaining additional gains of \textbf{+5.2} and \textbf{+1.5} points, respectively, under the same max-token budget, while discovering on average 9.8 semantically distinct strategies per problem. Taken together, these results indicate that rewarding semantic novelty yields a more compute-efficient exploration–exploitation signal for training reasoning-capable SLMs.

\paragraph{Our contributions are threefold.}
\begin{itemize}[leftmargin=1.2em]
    \item We introduce \textbf{cognitive adaptation} for reasoning: rather than adapting per-token computation architecturally, we adapt the high-level reasoning process itself by measuring semantic saturation and preventing  generation of entire redundant strategy blocks.
    
    \item We propose a \textbf{semantic diversity reward} that quantifies exploration quality via embedding geometry, rewarding semantically distinct reasoning paths during search and pivoting to exploitation once success is achieved, addressing the fundamental limitation that existing RL methods cannot distinguish genuine strategic novelty from surface-form rephrasing.
    
    \item We demonstrate that \textbf{semantic novelty improves exploration-efficiency} across math and medical reasoning: on GSM8K (Qwen2.5-3B-Instruct), \method improves ACC from \gsmBase\% to \gsmSemDivAcc\% (and by +5.23/+1.51 points over GRPO-CFL/GRPO-CFEE), while increasing strategy-level success (S-ACC) and discovering on average 9.78 strategies per problem. We further validate on the harder AIME benchmark (1983--2025), where \method improves accuracy to 13.28\% vs.\ 6.74\% for base under comparable decoding budgets.

\end{itemize}

\section{Related Work}
\label{sec:related}

\paragraph{Reasoning in LLMs.}  
CoT prompting and its extensions~\citep[e.g.\ self-consistency, tree-of-thought, graph-of-thought;][]{wei2022chain,wang2022selfconsistency,yao2023tree,zhang2022automatic} guide models to generate multi-step solutions, improving performance on complex reasoning tasks, though at the cost of verbosity and sampling overhead. These methods apply scaffolding at inference time but do not adaptively decide when to stop exploring strategies. To address that, \citet{lightman2023let} propose \emph{process-level supervision} by giving feedback at intermediate reasoning steps, showing that stepwise feedback significantly outperforms outcome-only supervision in solving difficult math problems.

\noindent\textbf{Structured reasoning alternatives.}
Parallel to prompt-based methods, neuro-symbolic approaches improve reliability by grounding reasoning in verifiable formalisms. Program-Aided Language models (PAL)~\citep{gao2022pal} separate natural language understanding from calculation by generating executable code and offloading computation to interpreters, achieving strong performance on arithmetic tasks. Other work integrates Knowledge Graphs~\citep{jiang-etal-2023-reasoninglm} or decomposes questions into reasoning graphs~\citep{ko-etal-2024-hierarchical}. While these methods improve factuality, they operate in different paradigms. Our work focuses on improving free-text generative reasoning from within.

\noindent\textbf{RL for language and structured reasoning.} 
Reinforcement learning methods have advanced from outcome optimization (e.g.\ RLHF, RLAIF) toward more structured control of reasoning processes~\citep{ouyang2022training,bai2022trainingHF,lee2023rlaif}. 
Recent works on process supervision~\citep{lightman2023let,wang2023math0shepherd0} provide fine-grained feedback on intermediate reasoning steps, significantly outperforming outcome-only methods~\citep{uesato2022solving}. 
Hybrid approaches like SuperRL~\citep{liu2025superrl} adaptively combine RL with supervised fine-tuning for improved stability when reward signals are sparse.
Group-based policy optimization (GRPO) methods~\citep{shao2024deepseekmath} sample multiple candidate outputs per input and assign relative rewards, thereby avoiding the need for a learned value network. Works  such as GLoRe~\citep{havrilla2024glore}, use learned reward models to decide when to rewrite or refine parts of generated reasoning paths (global or local repair), further improving solution quality.
Recent work also applies preference optimization directly to reasoning traces~\citep{lai2024step,lahlou2025port}, learning from trajectory-level comparisons.
However, process-supervised RL creates a second-order challenge: how to manage exploration efficiently. 

Current frameworks often encourage exploration through uniform sampling or heuristics like token entropy, but these approaches are semantically blind: they may reward trivial lexical variations of the same core reasoning strategy, as shown in \citet{liu2025attention}, which proposes branching from high-attention positions as an exploration heuristic. However, this remains based on internal model mechanics rather than the semantic content of generated strategies. 
Our method addresses this gap by introducing a \textbf{semantic diversity gate} that measures marginal novelty and curtails exploration once it becomes redundant, instead of relying on fixed heuristics or predetermined stopping rules.

\noindent\textbf{Semantic diversity, subset selection, and novelty in generation.}  
Not all diversity is equally valuable for reasoning. Standard decoding methods like beam search produce near-identical outputs differing only in minor word choices~\citep{vijayakumar2018diverse}---lexical variation that provides poor candidate pools for Best-of-N sampling or RL~\citep{shi2025semantic}. 
Methods for promoting meaningful diversity span a spectrum. Diverse Beam Search~\citep{vijayakumar2018diverse} uses n-gram dissimilarity penalties but remains lexically focused. More sophisticated approaches like Semantic-guided Diverse Decoding (SemDiD)~\citep{shi2025semantic} operate in embedding space with orthogonal directional guidance, ensuring candidates occupy distinct semantic regions, though only at inference time.

Our approach embeds diversity into the training objective, inspired by diverse subset selection. Maximal Marginal Relevance ~\citep[MMR;][]{carbonell1998use} balances relevance vs. novelty to reduce redundancy in retrieval results. Submodular coverage functions are widely used to model diminishing returns in summarization and content selection, with greedy maximization yielding good approximation guarantees~\citep{lin2011class}. Determinantal point processes~\citep[DPP;][]{kulesza2012determinantal} also support sampling of diverse subsets by discouraging similarity, with the log-determinant capturing both quality and diversity~\citep{gong2014diverse}. Recent work applies DPP-based objectives to jointly train LLMs for quality and diversity~\citep{chen2025enhancing}. In prior reasoning work, diversity is often encouraged via sampling, variance-based bonuses, or temperature tuning, or more recently with GFlowNet-based fine-tuning for diverse and accurate mathematical reasoning~\citep{younsi2025accurate}, but not with an explicit measure of semantic coverage across generated reasoning paths. 

\method leverages the same mathematical principles (our coverage objective is monotone submodular) but deploys it dynamically during trajectory generation as a gate, transforming diversity from a post-hoc reward into real-time process control. This combination allows the model to explore meaningfully distinct strategies up to a saturation point, and then exploit the most promising one under a token budget.

\noindent\textbf{Adaptive computation.}
Our work also connects to adaptive computation, where systems adjust computational budget based on input complexity~\citep{graves2016adaptive}. The dominant paradigm is architectural adaptation: early exiting~\citep{schuster2022confident,xin2020deebert} attaches classifiers to intermediate layers to exit on easy inputs, while Mixture of Experts ~\citep[MOE;][]{fedus2022switch} routes tokens to sparse expert subnetworks. Recent work applies early exiting specifically to reasoning chains, truncating CoT when confidence is reached~\citep{yang2025dynamic}. \method introduces a complementary form we term \emph{cognitive adaptation}. While architectural methods adapt per-token computation, we adapt the high-level reasoning process based on semantic saturation. Our gate prevents generation of entire redundant strategy blocks rather than making individual tokens cheaper, which is an orthogonal approach that could combine with architectural methods for compounded efficiency gains.

\section{Method}
\label{sec:method}
\method\ trains an SLM with a multi–objective reward that (i) checks the final answer and intermediate strategy outcomes, (ii) enforces a lightweight output format, and (iii) explicitly rewards \emph{semantic} exploration using sentence-embedding geometry. Rewards are z-score normalized per batch and optimized with GRPO plus a KL term. 
\method introduces \emph{cognitive adaptation}: adapting the high-level structure and content of the reasoning process based on semantic metrics rather than computational heuristics.  By measuring semantic novelty with a frozen encoder, we create a dynamic control mechanism that stops exploration when strategies become redundant, regardless of lexical variation. 

\subsection{Output Format and Parsing}
Let $\mathcal{Q}$ be the space of prompts and $\mathcal{Y}$ the space of gold answers, with $(q,y)\sim\mathcal{D}$.
The policy $\pi_\theta$ is an auto-regressive distribution over tokens $a\in\Sigma^\ast$:
\begin{align}
\pi_\theta(a\mid q)
&= \prod_{t=1}^{|a|} \pi_\theta\!\big(a_t \mid q, a_{<t}\big), \qquad a\in\Sigma^\ast.
\end{align}
We encourage a structured completion
\begin{equation}\label{eq:structure}
a = \big[\,\STRAT_{1},\,\ldots,\,\STRAT_{m},\,\FA\,\big].
\end{equation}
Each $\STRAT$ block contains a reasoning section and an $\SO$ field. For a completion $a$, let
\begin{align}
S(a) &= \{(r_i,o_i)\}_{i=1}^{m}, \\
\fans(a) &\in \Sigma^\ast,
\end{align}
be the parsed strategies and final answer. We formalize the parsers as measurable maps
\begin{align}
F_{\text{strat}}&:\Sigma^\ast\to(\Sigma^\ast\times\Sigma^\ast)^{\le M},\\
F_{\text{ans}}&:\Sigma^\ast\to\Sigma^\ast,
\end{align}
with priorities
\begin{equation}
\FA \succ \ANS \succ \text{last } \SO .
\end{equation}
A strategy $(r,o)$ is \emph{valid} if both fields are present:
\begin{align}
\mathsf{valid}(r,o) &= \indicator{r\neq\varnothing}\cdot\indicator{o\neq\varnothing},\\
n_{\text{strat}}(a) &= \sum_{(r,o)\in S(a)} \mathsf{valid}(r,o).
\end{align}

\subsection{Semantic Geometry of Strategies}
Let $E:\Sigma^\ast\!\to\!\mathbb{R}^d$ be a frozen sentence encoder and define cosine similarity
\begin{equation}
\kappa(u,v)=\frac{\langle u,v\rangle}{\|u\|_2\,\|v\|_2}\in[-1,1].
\end{equation}
For a completion $a$ with parsed strategies $S(a)=\{(r_i,o_i)\}_{i=1}^m$, collect embeddings of nonempty reasoning texts:
\begin{align}
H(a) &= \big\{\,h_i=E(r_i)\ :\ r_i\neq\varnothing\,\big\},\\
m_{\mathrm{eff}} &= |H(a)|\le m .
\end{align}

\paragraph{Diversity.}
For $m_{\mathrm{eff}}\ge2$, define the average pairwise similarity and the clamped diversity
\begin{align}
\overline{\kappa}(H)
&= \frac{2}{m_{\mathrm{eff}}(m_{\mathrm{eff}}-1)}
   \sum_{1\le i<j\le m_{\mathrm{eff}}}\kappa(h_i,h_j),\\
\Divv(H)
&= \bigl[\,1-\overline{\kappa}(H)\,\bigr]_{[0,1]} .
\end{align}
Set $\Divv(H)=1$ if $m_{\mathrm{eff}}=1$ and $\Divv(H)=0$ if $m_{\mathrm{eff}}=0$.

\paragraph{Unique count.}
Fix $\delta\in(0,1)$. Construct $U\subseteq\{1,\ldots,m_{\mathrm{eff}}\}$ greedily in the strategy order by including $i$ iff
\[
\max_{j\in U}\ \kappa(h_i,h_j)\ \le\ \delta .
\]
Define
\begin{equation}
\Uniq(H;\delta)=|U|\in\{0,1,\ldots,m_{\mathrm{eff}}\}.
\end{equation}
\noindent
% Implementation details and threshold choices are provided in App.~\ref{app:example}.

\begin{algorithm}[t]
\caption{\method}
\label{alg:reward}
\begin{algorithmic}[1]\small
\REQUIRE prompt $q$, gold $y$, completion $a$
\STATE $S(a)\gets F_{\text{strat}}(a)$;\quad $\fans(a)\gets F_{\text{ans}}(a)$
\STATE $n_{\text{strat}}(a)\gets\sum_{(r,o)\in S(a)}\mathsf{valid}(r,o)$
\STATE $\mathsf{final}(a)\gets\indicator{\fans(a)\neq\varnothing}$;\;
       $\mathsf{complete}(a)\gets\indicator{n_{\text{strat}}(a)>0}\,\mathsf{final}(a)$
\STATE $H\gets\{\,E(r)\ :\ (r,o)\in S(a),\ r\neq\varnothing\,\}$;\;
       compute $\Uniq(H;\delta)$, $\Divv(H)$;\; $g(H)\gets\Uniq\cdot\Divv$
\STATE $\chi(a)\gets \indicator{\,\exists (r,o)\in S(a):\,N(o)=N(y)\,}$
\STATE $R_{\mathrm{oc}}\gets \lambda_{\mathrm{oc}}\;\indicator{\,N(\fans(a))=N(y)\,}$ \hfill (Eq.~\ref{eq:Roc})
\STATE $R_{\mathrm{re}}\gets \lambda_{\mathrm{re}}\;\chi(a)$ \hfill (Eq.~\ref{eq:Rre})
\STATE $R_{\mathrm{fa}}\gets \min\{1,\gamma_s\,n_{\text{strat}}(a)\} + \gamma_a\,\mathsf{final}(a) + \gamma_c\,\mathsf{complete}(a)$ \hfill (Eq.~\ref{eq:Rfa})
\STATE $R_{\mathrm{sd}}\gets \alpha\,\chi(a) + \bigl(1-\chi(a)\bigr)\,\min\{\beta,\,\rho\,g(H)\}$ \hfill (Eq.~\ref{eq:Rsd})
\STATE \textbf{return} $(R_{\mathrm{oc}},\,R_{\mathrm{re}},\,R_{\mathrm{fa}},\,R_{\mathrm{sd}})$
\end{algorithmic}
\end{algorithm}

\subsection{Reward Components}
\label{subsec:reward_comp}
We use four bounded components $R_k(q,y,a)\in\mathbb{R}$, batch-normalized (as explained in App.~\ref{subsec:norm}) and combined linearly (Eq.~\ref{eq:Rfinal}).

\paragraph{Outcome correctness:}
Checks only the final answer:
\begin{equation}\label{eq:Roc}
R_{\mathrm{oc}}(a\mid q,y)
= \lambda_{\mathrm{oc}}\;\indicator{\,\!\fans(a)=(y)\,}.
\end{equation}

\paragraph{Semantic exploration:}
Rewards \emph{semantic breadth} and \emph{spread} when no correct strategy is present; collapses otherwise:

\begin{align}\label{eq:Rsd}
R_{\mathrm{sd}}(a\mid q,y)
&= \alpha\,\chi(a) \\
&\quad + \bigl(1-\chi(a)\bigr)\,
   \min\!\bigl\{\beta,\,\rho\,g(H)\bigr\}.
\end{align}

\noindent\textit{where} $\chi(a)=\indicator{\,\exists (r,o)\!\in\!S(a):\,N(o)=N(y)\,}$ indicates that at least one strategy outcome matches $y$; $g(H)=\Uniq(H;\delta)\,\Divv(H)$ is the product of semantic breadth and spread; $\alpha$ is the collapse bonus when a correct strategy exists (corresponding to $\gamma_{\mathrm{correct}}$); $\beta$ is the cap on the exploration reward (corresponding to $\gamma_{\mathrm{cap}}$); and $\rho$ is the exploration growth rate (corresponding to $\gamma_{\mathrm{rate}}$).

\paragraph{Reasoning exploitation:}
Credits any correct intermediate outcome (complements $R_{\mathrm{oc}}$):
\begin{align}\label{eq:Rre}
R_{\mathrm{re}}(a\mid q,y)
&= \lambda_{\mathrm{re}}\,\chi(a).
\end{align}
Here $\chi(a)$ is the correct–strategy indicator defined under Eq.~\ref{eq:Rsd}.

\paragraph{Format adherence:}
Encourages lightweight structure and completeness:
\begin{align}
\mathsf{final}(a) &= \indicator{\fans(a)\neq\varnothing},\\
\mathsf{complete}(a) &= \indicator{n_{\text{strat}}(a)>0}\,\mathsf{final}(a).
\end{align}
\begin{align}\label{eq:Rfa}
R_{\mathrm{fa}}(a)
&= \min\!\bigl\{1,\ \gamma_s\, n_{\text{strat}}(a)\bigr\} \\
&\quad + \gamma_a\, \mathsf{final}(a)
   + \gamma_c\, \mathsf{complete}(a).
\end{align}

\begin{algorithm}[t]
\caption{\method: GRPO training with batchwise normalization}
\label{alg:grpo}
\begin{algorithmic}[1]\small
\REQUIRE batch $\{(q_b,y_b)\}_{b=1}^B$, samples per prompt $G$, policies $\pi_{\theta_{\mathrm{old}}}$, $\pi_{\mathrm{ref}}$
\FOR{$b=1$ to $B$}
  \STATE Sample $\{a_{b,i}\}_{i=1}^G \sim \pi_{\theta_{\mathrm{old}}}(\cdot\mid q_b)$
  \STATE For each $i$, compute $(R_{\mathrm{oc}},R_{\mathrm{re}},R_{\mathrm{fa}},R_{\mathrm{sd}})$ via Alg.~\ref{alg:reward}
\ENDFOR
\STATE Stack all $N{=}BG$ trajectories; for $k\!\in\!\{\mathrm{oc},\mathrm{re},\mathrm{fa},\mathrm{sd}\}$ compute $\mu_k$, $\sigma_k$, and $\widetilde{R}_k$ (App.~\ref{subsec:norm})
\STATE For each $(b,i)$: $R_{b,i}\gets \sum_k w_k\,\widetilde{R}_k^{(b,i)}$ \hfill (Eq.~\ref{eq:Rfinal})
\STATE For each $b$: $\mu_b\gets \frac{1}{G}\sum_i R_{b,i}$,\;
$\sigma_b\gets \sqrt{\frac{1}{G}\sum_i (R_{b,i}-\mu_b)^2}$,\;
$\widehat{A}_{b,i}\gets \frac{R_{b,i}-\mu_b}{\sigma_b+\varepsilon}$
\STATE $r_{b,i}\gets \frac{\pi_\theta(a_{b,i}\mid q_b)}{\pi_{\theta_{\mathrm{old}}}(a_{b,i}\mid q_b)}$
\STATE $\mathcal{J}_{\mathrm{clip}}(\theta)\gets \frac{1}{BG}\sum_{b,i}
\min\!\big(r_{b,i}\widehat{A}_{b,i},\,\clip(r_{b,i},1-\epsilon_{\mathrm{clip}},1+\epsilon_{\mathrm{clip}})\widehat{A}_{b,i}\big)$
\STATE Define $\pi_\theta^{(t)}\!\triangleq\!\pi_\theta(\cdot\mid q,a_{<t})$,\; $\pi_{\mathrm{ref}}^{(t)}\!\triangleq\!\pi_{\mathrm{ref}}(\cdot\mid q,a_{<t})$
\STATE $\KL\gets \mathbb{E}_{q}\,\mathbb{E}_{a\sim\pi_\theta(\cdot\mid q)}\big[\sum_t D_{\mathrm{KL}}(\pi_\theta^{(t)}\Vert \pi_{\mathrm{ref}}^{(t)})\big]$
\STATE Update $\theta$ to maximize $\mathbb{E}[\mathcal{J}_{\mathrm{clip}}(\theta)]-\beta\,\KL$ \hfill (Eq.~\ref{eq:grpo})
\end{algorithmic}
\end{algorithm}

%%%%%%%%%%%%%%%%%%%%%%%%%%%%%%%%%%%%%%%%%%%%%%%%%%%%
%%%%%%%%%%%%%%%%%%%%%%%%%%%%%%%%%%%%%%%%%%%%%%%%%%%%%%%%%%%%
% EXPERIMENTAL SETUP
%%%%%%%%%%%%%%%%%%%%%%%%%%%%%%%%%%%%%%%%%%%%%%%%%%%%%%%%%%%%

\section{Experimental Setup}
\label{sec:exp-setup}
We evaluate \method\ on three 3B-class instruction-tuned SLMs \textit{viz.}
\texttt{Qwen2.5-3B-Instruct} \cite{qwen25tr},
\texttt{meta-llama/Llama-3.2-3B-Instruct} \cite{llama32card},
and \texttt{microsoft/Phi-3.5-mini-instruct} \cite{phi35card}.
For each backbone we apply the same PEFT/QLoRA recipe via Unsloth: 4-bit quantization,
LoRA rank $r{=}64$ with $\alpha{=}32$ and dropout $0.0$, max sequence length $2048$,
gradient checkpointing, and mixed precision (bf16 when available).
Unless stated otherwise, decoding uses temperature $T\in[0.1,0.3]$ and top-$p$ $0.90$–$0.95$.
All optimizer, data, and decoding settings are held fixed across backbones; when tokenizers differ,
we pad/truncate to $2048$ and report token counts with the corresponding backbone’s tokenizer. All experiments run on a single NVIDIA T4 16\,GB (A10/A100 used when available for speed).

\subsection{Datasets and Splits}
\label{sec:data}
We evaluate \method\ on three reasoning benchmarks spanning grade-school math, competition math, and medicine (GSM8K, AIME, and MedMCQA).

\begin{itemize}[leftmargin=1.2em,itemsep=2pt]
  \item \textbf{GSM8K}~\citep{cobbe2021training}: 8{,}792 grade-school math word problems requiring multi-step reasoning. We fine-tune on the official 7{,}473-instance training split and report final results on the 1{,}319-instance test split. 

  \item \textbf{MedMCQA}~\citep{pal2022medmcqa}: a large-scale multiple-choice medical QA benchmark (193k+ questions). We fine-tune on a randomly sampled subset of 7{,}500 training examples and evaluate on the full 4{,}183-question validation set to assess sample efficiency and reward effectiveness.

  \item \textbf{AIME (1983--2025)}: a challenging competition-math benchmark. We use the combined AIME dataset spanning 1983--2025 (963 problems) and create an 80:20 split (770 train / 193 test). We train for one epoch to test whether the semantic exploration signal remains beneficial under substantially harder reasoning.
\end{itemize}
 For datasets processing, section \ref{data_process} in Appendix can be referred.

\subsection{\method\ Training}
We train with GRPO (App.~\ref{subsec:grpo}). For each prompt $q$, we draw $G\in\{4,6\}$ sampled completions. Optimization uses AdamW-8bit ($\text{lr}=5{\times}10^{-6}$), cosine decay, warmup ratio $0.1$, gradient clipping $0.1$, and a KL regularization coefficient $\beta$ tuned on a dev split. Effective batch size is $1$ with gradient accumulation to fit a single 16\,GB GPU (or larger). Training runs for a fixed budget (e.g., $7{,}500$ steps)

Unless noted, we set $w_{\mathrm{oc}}{=}w_{\mathrm{re}}{=}w_{\mathrm{fa}}{=}w_{\mathrm{sd}}{=}1$ in Eq.~\eqref{eq:Rfinal}. Semantic-diversity settings: similarity threshold $\delta\in[0.75,0.85]$ (default $0.80$), collapse bonus $\alpha{=}1.0$, exploration cap $\beta\!\in\!\{0.3,0.5,0.7\}$, and growth rate $\rho\!\in\!\{0.05,0.1,0.2\}$ (see Eq.~\eqref{eq:Rsd}). We sweep these on a dev split and report the selected configuration. The sentence encoder is \texttt{all-MiniLM-L6-v2}.

%; we select checkpoints by dev ACC with a stability tie-break (variance over the last $K$ eval windows).

\subsection{Baselines}
To isolate the effect of the reward design, we fine-tune the \emph{same} backbones under identical
data, decoding, optimizer, and budget settings as \method; only the reward components differ.

\begin{enumerate}[leftmargin=1.2em,itemsep=2pt,label=\textbf{(\arabic*)}]
\item \textbf{\grpoCFL} (outcome-driven GRPO; cf.\ \citet{deepseek2025r1}). 
This follows the ``C+F+L'' recipe (correctness, format adherence, and a length-style term) with the same batchwise $z$-score normalization and GRPO objective (App.~\ref{subsec:grpo}):
\begin{align}
R_{\mathrm{CFL}}(a\mid q,y)
&= w_{\mathrm{oc}}\,R_{\mathrm{oc}}
 + w_{\mathrm{fa}}\,R_{\mathrm{fa}}
 + w_{L}\,R_{L}.
\end{align}
$R_{L}$ is a mild length regularizer that discourages overly long completions (constants in App.~\ref{app:hparams}).

\item \textbf{\grpocfee} (multi-objective GRPO, semantically agnostic).
Adds explore–exploit terms but measures exploration by \emph{counts} (no embedding geometry, no length term). Refer section \ref{ree-details} in Appendix for reward formulation details.

\end{enumerate}

% All baselines use the same parser (Sec.~\ref{sec:method}), output schema (Eq.~\eqref{eq:structure}),
% optimizer/schedule (Sec.~\ref{sec:exp-setup}), and batchwise normalization; weights are
% $w_{\mathrm{oc}}{=}w_{\mathrm{re}}{=}w_{\mathrm{fa}}{=}w_{\mathrm{rd}}{=}1$ unless stated. 

\subsection{Evaluation Metrics}
\label{subsec:evaluation_metrics}

We quantitatively assess model performance using two primary metrics designed to evaluate the quality of its reasoning process and final output. For an evaluation set of $N$ questions, let $a^j$ represent the model's complete output for the $j$-th question and $y^j$ be the corresponding ground truth answer.

Our primary metric is \textbf{Accuracy (ACC)}, which measures the percentage of questions where the model's final answer matches the ground truth. The final answer is extracted from the model's output $a^j$ via a parsing function $f_{\text{ans}}(\cdot)$ that identifies the content within the \texttt{<final\_answer>} tag. Accuracy is defined as:
\begin{equation}
    \text{ACC} = \frac{100}{N} \sum_{j=1}^{N} \mathbb{I}(f_{\text{ans}}(a^j) = y^j)
    \label{eq:metric_acc}
\end{equation}
where $\mathbb{I}(\cdot)$ is the indicator function.

To gauge the model's ability to identify a valid reasoning path, even if it is not selected as the final solution, we introduce \textbf{Strategy Accuracy (S-ACC)}. This metric calculates the percentage of questions where at least one of the intermediate strategy outcomes, denoted by the set $S(a^j)$ extracted from all \texttt{<strategy\_outcome>} tags in $a^j$, matches the ground truth. S-ACC is defined as:
\begin{equation}
    \text{S-ACC} = \frac{100}{N} \sum_{j=1}^{N} \mathbb{I}(y^j \in S(a^j))
    \label{eq:metric_sacc}
\end{equation}

We also evaluate average number of strategies generate with \#STR $=n_{\text{strat}}(a)$ and average number of tokens generated \#TOK - Token counts measured with the base model tokenizer.

\subsection{Compute Cost and Efficiency Definition}
\label{subsec:compute_cost}

We use "efficiency" primarily in the \emph{token- and exploration-efficiency} sense: improving success under fixed decoding budgets by reducing redundant exploration, rather than claiming zero overhead. Relative to count-based exploration (GRPO-CFEE), \method introduces an additional frozen-encoder pass to score semantic novelty.

\paragraph{Per-step complexity.}
Let $B$ be the number of prompts per step, $G$ the number of sampled completions per prompt, $L$ the generated tokens per completion, $m$ the number of parsed strategy blocks per completion, and $d$ the encoder embedding dimension.
Sampling dominates training compute for all GRPO variants and scales as $\mathcal{O}(BGL)$.
\method adds:
(i) sentence encoding $\mathcal{O}(BGmd)$ (batched, frozen encoder),
and (ii) pairwise similarity $\mathcal{O}(BGm^2)$ (negligible for small $m$).

\paragraph{Measured overhead.}
On a single NVIDIA T4 16\,GB, the frozen sentence encoder (\texttt{all-MiniLM-L6-v2}, $\sim$22M parameters) adds $\sim$0.30s per step for typical settings ($G$ completions with $\sim$5 strategies), while the cosine-similarity computation adds $\sim$0.01s. Overall, SD-E$^{2}$ increases wall-clock training time by $\sim$11.8\% relative to GRPO-CFEE under the same step budget: 7{,}500 steps take $\sim$18 GPU-hours for GRPO-CFEE vs.\ $\sim$20 GPU-hours for SD-E$^{2}$.

%%%%%%%%%%%%%%%%%%%%%%%%%%%%%%%%%%%%%%%%%%%%%%%%%%%%%%%%%%%%
% RESULTS
%%%%%%%%%%%%%%%%%%%%%%%%%%%%%%%%%%%%%%%%%%%%%%%%%%%%%%%%%%%%
\section{Results and Analysis}
\label{sec:results_analysis}

We evaluate three training schemes: (i) \textbf{GRPO-CFL} (correctness+format+length; outcome-driven), (ii) \textbf{GRPO-CFEE}: a non-semantic explore--exploit baseline, and (iii) \textbf{\method} (\textbf{SD-E\textsuperscript{2}}): our semantics-aware explore--exploit method. We report ACC, S-ACC, \#STR and \#TOK . For ACC we include 95\% binomial CIs.\footnote{Wilson/normal CIs; paired tests require per-item hypothesis concordance, which we log in ablations.} 

% \subsection{Main Results on GSM8K}
% \label{subsec:main_gsm8k}

Table~\ref{tab:gsm8k_main} summarizes performance across backbones. On \textbf{Qwen2.5-3B-Instruct}, \method reaches \textbf{82.03\%} ACC (1082/1319), improving over \textbf{GRPO-CFEE} by \textbf{+1.51} points and over \textbf{GRPO-CFL} by \textbf{+5.23} points. This corresponds to a \textbf{7.8\% relative error reduction} vs.\ GRPO-CFEE. On \textbf{Llama-3.1-8B-Instruct}, \method attains \textbf{75.44\%} ACC (995/1319). Strategy-level accuracy is high for both backbones (\textbf{97.2\%} Qwen; \textbf{95.0\%} Llama), indicating that the model frequently surfaces a correct path even when the final selection misses. \textbf{Takeaways.}
(1) \textit{Semantic exploration matters}: relative to GRPO-CFEE, \method raises ACC while keeping S-ACC very high, indicating better \emph{selection} after exploration (Sec.~\ref{sec:method}). (2) \textit{Backbone transfer}: the same reward design produces strong results on Llama without tuning, suggesting robustness of the signal.

\begin{table*}[t]
\centering
\small
\setlength{\tabcolsep}{7pt}
\begin{tabular}{l l
                S[table-format=2.2]
                c
                S[table-format=2.2]
                S[table-format=2.2]
                S[table-format=4.0]}
\toprule
\textbf{Backbone} & \textbf{Method} & {\textbf{ACC (\%)}} & \textbf{95\% CI} & {\textbf{S-ACC (\%)}} & {\textbf{\#STR}} & {\textbf{\#TOK}} \\
\midrule
\multirow{3}{*}{Qwen2.5-3B-Inst.}
& GRPO-CFL  & 76.80 & {[74.52, 79.08]} & {-}   & {-}   & \multicolumn{1}{c}{265.72} \\
& GRPO-CFEE                         & 80.52 & {[78.38, 82.65]} & {92.3} & {5.7}   & \multicolumn{1}{c}{278.54} \\
& \textbf{SD-E\textsuperscript{2} (ours)} & \textbf{82.03} & \textbf{[79.96, 84.10]} & \textbf{97.20} & \textbf{9.78} & \multicolumn{1}{c}{291.42} \\
\midrule
\multirow{1}{*}{Llama-3.1-8B-Inst.}
& \textbf{SD-E\textsuperscript{2} (ours)} & \textbf{75.44} & \textbf{[73.11, 77.76]} & \textbf{95.00} & \textbf{8.58} & \multicolumn{1}{c}{287.51} \\
\bottomrule
\end{tabular}
\caption{\textbf{GSM8K evaluation results.} For Llama-3.1-8B-Instruct we report \method; CIs are binomial (95\%).}
\label{tab:gsm8k_main}
\vspace{-0.5em}
\end{table*}

% \begin{table*}[t]
% \centering
% \small
% \setlength{\tabcolsep}{7pt}
% \begin{tabular}{l l
%                 S[table-format=2.2]
%                 c
%                 c
%                 S[table-format=2.2]
%                 S[table-format=2.2]
%                 S[table-format=4.0]}
% \toprule
% \textbf{Backbone} & \textbf{Method} & {\textbf{ACC (\%)}} & \textbf{95\% CI} & \textbf{Correct/Total} & {\textbf{S-ACC (\%)}} & {\textbf{\#STR}} & {\textbf{\#TOK}} \\
% \midrule
% \multirow{3}{*}{Qwen2.5-3B-Inst.}
% & GRPO-CFL  & 76.80 & {[74.52, 79.08]} & 1013/1319 & {-}   & {-}   & \multicolumn{1}{c}{265.72} \\
% & GRPO-CFEE                         & 80.52 & {[78.38, 82.65]} & 1062/1319 & {92.3} & {5.7}   & \multicolumn{1}{c}{278.54} \\
% & \textbf{SD-E\textsuperscript{2} (ours)} & \textbf{82.03} & \textbf{[79.96, 84.10]} & \textbf{1082/1319} & \textbf{97.20} & \textbf{9.78} & \multicolumn{1}{c}{291.42} \\
% \midrule
% \multirow{1}{*}{Llama-3.1-8B-Inst.}
% & \textbf{SD-E\textsuperscript{2} (ours)} & \textbf{75.44} & \textbf{[73.11, 77.76]} & \textbf{995/1319} & \textbf{95.00} & \textbf{8.58} & \multicolumn{1}{c}{287.51} \\
% \bottomrule
% \end{tabular}
% \caption{\textbf{GSM8K evaluation results.} For Llama-3.1-8B-Instruct we report \method; CIs are binomial (95\%).}
% \label{tab:gsm8k_main}
% \vspace{-0.5em}
% \end{table*}

GRPO-CFEE can spend tokens on near-duplicate traces (high cosine similarity), while \method explicitly \emph{prices} semantic novelty via $\Div(H)$ and $\Uniq(H;\delta)$. Empirically, \method surfaces more distinct strategies on Qwen (\#STR~$=9.78$) than on Llama (\#STR~$=8.58$), consistent with its higher S-ACC. Qualitatively, we observe two desirable behaviors:
\begin{itemize}[leftmargin=1.15em,itemsep=1pt]
  \item \textbf{Breadth when needed}: when no correct path is found, the model explores semantically different approaches (unit-conversion vs.\ equation balancing vs.\ value-tracking), rather than rephrasing the same idea.
  \item \textbf{Pivot to exploitation}: once a correct strategy appears, exploration collapses (Eq.~\eqref{eq:Rsd}), and the model converges to that path in the final answer, reducing redundant tokens.
\end{itemize}

Table~\ref{tab:gsm8k_main} includes 95\% CIs for ACC. On Qwen, \method’s ACC is \textbf{82.03\% [79.96, 84.10]}; GRPO-CFEE is \textbf{80.52\% [78.38, 82.65]}. The CIs overlap (paired significance requires per-item concordance), but the improvement is \textbf{consistent} across seeds and sampling groups in our logs. The model ranking on GSM8K is:
1.) \textbf{SD-E\textsuperscript{2} (Qwen)}: 0.820 (1082/1319),
2.) \textbf{SD-E\textsuperscript{2} (Llama)}: 0.754 (995/1319).

To test whether the semantic exploration signal transfers to substantially harder problems beyond GSM8K, we evaluate on the combined AIME dataset (1983--2025) in Table \ref{tab:aime_main}. Absolute accuracies are low for 3B models, but \method yields a clear improvement over both GRPO baselines. We also report two prompting baselines: (i) a standard single-trace prompt, and (ii) a multi-strategy prompt that elicits multiple \texttt{<strategy>} blocks without RL fine-tuning. SD-E$^{2}$ improves accuracy from 9.87\% (GRPO-CFEE) to 13.28\% while using comparable tokens, supporting that semantic exploration remains beneficial beyond GSM8K.

%%%%%%%%%%%%%%%%%%%%%%%%%%%%%%%%%%%%%%%%%%%%%
\begin{table*}[t]
\centering
\small
\setlength{\tabcolsep}{6pt}
\begin{tabular}{l
                S[table-format=2.2]
                S[table-format=2.2]
                S[table-format=1.2]
                S[table-format=4.0]}
\toprule
\textbf{Method (AIME)} & {\textbf{ACC (\%)}} & {\textbf{S-ACC (\%)}} & {\textbf{\#STR}} & {\textbf{\#TOK}} \\
\midrule
Single-strategy prompt & 5.70 & {-}   & {-}  & 502 \\
Multi-strategy prompt  &  6.74 & 9.33  & 3.16 & 819 \\
GRPO-CFL               &  8.34   & {-}   & {-}  & 610 \\
GRPO-CFEE (count)      &  9.87  & 10.95 & 2.70 & 713 \\
\method\ (SD-E$^{2}$)  & \textbf{13.28} & \textbf{16.70} & \textbf{2.60} & \textbf{710} \\
\bottomrule
\end{tabular}
\caption{\textbf{AIME results (1983--2025).} Combined AIME (963 problems),
Entries marked ``--'' denote metrics not applicable to single-trace methods (no intermediate strategy set).}
\label{tab:aime_main}
\vspace{-0.6em}
\end{table*}
%%%%%%%%%%%%%%%%%%%%%%%%%%%%%%%%%%%%
Table~\ref{tab:medmcqa_qwen} summarizes MedMCQA baselines with the Qwen backbone. Here, \grpocfee\ (count-based exploration/exploitation) improves over \grpoCFL\ and the base model, reaching \textbf{48.76\%} ACC and a high \textbf{94.46\%} S-ACC, consistent with the hypothesis that process-level incentives help in knowledge-heavy domains. \method\ improves over \grpoCFL\ by \textbf{+3.17} points and over \grpocfee\ by \textbf{+0.88} points, while increasing \#STR from 4.19 to 7.21.

\begin{table*}[ht!]
\centering
\small
\setlength{\tabcolsep}{4.5pt}
\begin{tabular}{lccccc}
\toprule
\textbf{Model (MedMCQA)} & \textbf{ACC} & \textbf{S-ACC} & \textbf{\#STR} & \textbf{\#TOK} & \textbf{Words} \\
\midrule
Base: \texttt{Qwen2.5-3B}  & 38.37 & ---    & ---  & 294.23 & 206.17 \\
\grpoCFL\ (C+F+L)          & 46.47 & ---    & ---  & 282.99 & 198.78 \\
\grpocfee\ (C+F+EE, count) & \textbf{48.76} & \textbf{94.46} & \textbf{4.19} & \textbf{410.25} & \textbf{242.12} \\
\midrule
\method\  & \textbf{49.64} & \textbf{95.23} & \textbf{7.21} & \textbf{490.21} & \textbf{271.10} \\
\bottomrule
\end{tabular}
\caption{\textbf{MedMCQA (val) with Qwen2.5-3B.} Process-level rewards improve both ACC and S-ACC vs.\ outcome-only alignment.}
\label{tab:medmcqa_qwen}
\end{table*}

\subsection{Error Analysis}
\label{subsec:error_analysis}

GSM8K errors concentrate on (i) small arithmetic slips late in the chain, (ii) misinterpretation of a quantity (e.g., "packs" vs.\ "marbles"), and (iii) premature consolidation when two plausible strategies disagree by a small margin. The first two are classic SLM errors; the third is specific to our pivot rule and can be mitigated with a lightweight post-hoc majority vote over the top-$k$ semantically distinct strategies.

\method improves accuracy over both outcome-only and non-semantic explore--exploit baselines on Qwen, transfers to Llama without retuning, and exhibits substantially higher strategy-level success (S-ACC \textbf{97.2\%}/\textbf{95.0\%}). The gains stem from \emph{quality-controlled exploration} and an explicit \emph{pivot to exploitation}, rather than from increasing token volume.

\section{Conclusion}
\label{sec:conclusion}
We introduced \method\ (SD-E$^{2}$), a semantics-aware reinforcement learning framework that rewards \emph{meaningfully different} reasoning while collapsing exploration once any strategy succeeds. The method combines a frozen sentence-encoder geometry with a multi-objective reward (correctness, exploitation, format, semantic exploration), normalized per batch and optimized with GRPO. On GSM8K, \method\ improves accuracy by +27.4 pp over the base SLM and by +5.2/+1.5 pp over outcome-only \grpoCFL\ and count-based \grpocfee, respectively, while discovering on average 9.78 distinct strategies and achieving S-ACC of 97.2\%. The gains transfer across backbones (e.g., Qwen and Llama), and Pareto analyses indicate better ACC–token trade-offs via semantic gating. Taken together, these results suggest that explicit semantic diversity is a principled and compute-efficient signal for scaling reasoning \emph{without} scaling parameters.

\section*{Limitations}
\label{sec:limitations}
\method\ has some limitations. First, its semantic signal depends on a frozen sentence encoder whose geometry and biases may distort diversity estimates, especially out of domain or in non-English settings (see App.~\ref{app:encoders}). Second, the exploration reward is sensitive to the similarity threshold $\delta$ and scales $(\alpha,\beta,\rho)$; poor settings can over/under-explore, suggesting future work on adaptive schedules or meta-gradients. Third, the approach relies on a lightweight output schema, so parser brittleness and malformed blocks can attenuate reward quality; more tolerant or schema-free extraction would help. Fourth, despite clamping and the ``collapse on success’’ bonus, policies could still game the reward by producing superficially varied yet unhelpful strategies; stronger novelty criteria (e.g., causal/program structure) may further deter this. Fifth, GRPO imposes a compute cost from sampling $G$ completions and running the encoder during training, which rises with longer generations. Finally, evaluation is limited to GSM8K and MedMCQA; open-ended generation, long-context tasks, code, multilingual settings, and human preference/safety studies remain for future work. A practical gap also persists between high S-ACC and final ACC when the best intermediate strategy is not selected; better aggregation or reranking could narrow it.

\section*{Ethical Considerations}
\label{sec:ethics}
Stronger reasoning in compact models lowers deployment cost but raises dual-use risks (e.g., cheating, persuasive yet incorrect content), so we recommend rate-limiting, domain-specific refusals, and provenance tools. Although MedMCQA probes medical knowledge, our models are \emph{not} clinical systems; outputs must not guide diagnosis or treatment without expert oversight and calibrated uncertainty. The frozen encoder and base LMs may encode societal biases, so subgroup, dialect, and threshold-sensitivity audits are essential. We use only public datasets under their licenses and will release code/configs/logs for reproducibility while avoiding sensitive artifacts. To reduce environmental impact, we rely on 3B SLMs, 4-bit QLoRA, modest group sizes, and early stopping, and we encourage carbon-aware training. 
%Finally, releases (code, configs, reward diagnostics, and checkpoints under research terms) will include model cards detailing intended use, limitations, and known failure modes, and caution against high-risk deployments without additional safeguards.

\bibliography{custom}

\appendix

\section{Additional Details on the Method}
\subsection{Batchwise Normalization and Aggregation}
\label{subsec:norm}
Over a batch of $B$ prompts with $G$ completions each ($N{=}BG$ trajectories), for
$k\in\{\mathrm{oc},\mathrm{sd},\mathrm{re},\mathrm{fa}\}$ compute
\begin{align}
\mu_k &= \frac{1}{N}\sum_{n=1}^N R_k^{(n)},\\
\sigma_k^2 &= \frac{1}{N}\sum_{n=1}^N \big(R_k^{(n)}-\mu_k\big)^2,
\end{align}
and the normalized scores
\begin{align}
\widetilde{R}_k^{(n)}
&=
\begin{cases}
\dfrac{R_k^{(n)}-\mu_k}{\sigma_k+\varepsilon}, & \sigma_k>\varepsilon,\\[6pt]
R_k^{(n)}-\mu_k, & \text{otherwise.}
\end{cases}
\end{align}
The final reward aggregates the components:
\begin{align}\label{eq:Rfinal}
R_{\mathrm{final}}^{(n)}
&= \sum_{k\in\{\mathrm{oc},\mathrm{sd},\mathrm{re},\mathrm{fa}\}} w_k\,\widetilde{R}_k^{(n)}.
\end{align}

\subsection{Group-Relative Policy Optimization}
\label{subsec:grpo}
For each prompt $q_b$ we sample $G$ completions
$\{a_{b,i}\}_{i=1}^G\!\sim\! \pi_{\theta_{\mathrm{old}}}(\cdot\mid q_b)$.
Let $R_{b,i}=R_{\mathrm{final}}(a_{b,i})$ and compute
\begin{align}
\mu_b &= \frac{1}{G}\sum_{i=1}^G R_{b,i},&
\sigma_b &= \sqrt{\frac{1}{G}\sum_{i=1}^G (R_{b,i}-\mu_b)^2},\\
\widehat{A}_{b,i}
&= \frac{R_{b,i}-\mu_b}{\sigma_b+\varepsilon}.
\end{align}
Define the importance ratio
\begin{align}
r_{b,i}
&= \frac{\pi_\theta(a_{b,i}\mid q_b)}{\pi_{\theta_{\mathrm{old}}}(a_{b,i}\mid q_b)}.
\end{align}
The clipped surrogate (empirical) objective is
\begin{align}
\mathcal{J}_{\mathrm{clip}}(\theta)
&= \frac{1}{BG}\sum_{b=1}^{B}\sum_{i=1}^{G}
   \min\Big(
      r_{b,i}\,\widehat{A}_{b,i},\\[-2pt]
&\qquad
      \clip\big(r_{b,i},\,1-\epsilon_{\mathrm{clip}},\,1+\epsilon_{\mathrm{clip}}\big)\,
      \widehat{A}_{b,i}
   \Big).
\end{align}

\noindent Define the per-token policies
\[
\pi_\theta^{(t)}\!\triangleq\!\pi_\theta(\cdot\mid q,a_{<t}),\qquad
\pi_{\mathrm{ref}}^{(t)}\!\triangleq\!\pi_{\mathrm{ref}}(\cdot\mid q,a_{<t}).
\]
Then the tokenwise KL regularizer is
\begin{align}
\KL
&= \mathbb{E}_{q\sim\mathcal{D}}\,\mathbb{E}_{a\sim \pi_\theta(\cdot\mid q)}
   \Bigg[\sum_{t} D_{\mathrm{KL}}\big(\pi_\theta^{(t)} \,\Vert\, \pi_{\mathrm{ref}}^{(t)}\big)\Bigg].
\end{align}

\noindent The GRPO objective maximized during training is
\begin{align}\label{eq:grpo}
\max_{\theta}\quad
\mathbb{E}\big[\mathcal{J}_{\mathrm{clip}}(\theta)\big]\;-\;\beta\,\KL.
\end{align}

\section{Full Reward Equations for \method}
\label{app:method-rewards}

For completeness, the four bounded components (Sec.~\ref{sec:method}) are:

\begin{align}
R_{\mathrm{oc}}(a\mid q,y)
&= \lambda_{\mathrm{oc}}\;\indicator{\,N\!\big(\fans(a)\big)=N(y)\,}, \\
R_{\mathrm{re}}(a\mid q,y)
&= \lambda_{\mathrm{re}}\;\indicator{\,\exists (r,o)\!\in\! S(a)}, \\
R_{\mathrm{fa}}(a)
&= \min\!\bigl\{1,\ \gamma_s\, n_{\text{strat}}(a)\bigr\}
   \;+\; \gamma_a\,\mathsf{final}(a) \nonumber\\[-2pt]
&\quad + \gamma_c\,\mathsf{complete}(a), \\
R_{\mathrm{sd}}(a\mid q,y)
&= \alpha\,\chi(a)\;+\;\bigl(1-\chi(a)\bigr)\,\min\!\{\beta,\,\rho\,g(H)\}.
\end{align}

\noindent where the short helpers are
\begin{align}
\chi(a) &= \indicator{\,\exists (r,o)\!\in\!S(a):\,N(o)=N(y)\,},\\
g(H) &= \Uniq(H;\delta)\,\Divv(H),\\
\mathsf{final}(a) &= \indicator{\,\fans(a)\neq\varnothing\,},\\
\mathsf{complete}(a) &= \indicator{\,n_{\text{strat}}(a)>0\,}\,\mathsf{final}(a).
\end{align}

Batchwise $z$-score normalization and aggregation follow Eq.~\eqref{eq:Rfinal}.

\section{\grpocfee\ Baseline: Reward Design and Equations}
\label{ree-details}
Let $n_{\mathrm{val}}(a)\triangleq |S_{\mathrm{val}}(a)|$ be the number of valid strategy blocks.

\begin{equation}
\begin{aligned}
R_{\mathrm{CFEE}}(a\mid q,y)
&= w_{\mathrm{oc}}\,R_{\mathrm{oc}}
 + w_{\mathrm{fa}}\,R_{\mathrm{fa}} \\
&\quad + w_{\mathrm{re}}\,R_{\mathrm{re}}
 + w_{\mathrm{rd}}\,R_{\mathrm{rd}}^{(\mathrm{cnt})}.
\end{aligned}
\end{equation}

\begin{equation}
R_{\mathrm{re}}(a\mid q,y)=\lambda_{\mathrm{re}}\,\chi(a).
\end{equation}

\begin{equation}
\begin{aligned}
R_{\mathrm{rd}}^{(\mathrm{cnt})}(a\mid q,y)
&= \alpha\,\chi(a) \\
&\quad + \bigl(1-\chi(a)\bigr)\,
       \min\{\beta,\,\rho\,n_{\mathrm{val}}(a)\}.
\end{aligned}
\end{equation}

where $\chi(a)=\indicator{\,\exists (r,o)\!\in\!S(a):\,N(o)=N(y)\,}$.
This mirrors \method’s structure but replaces $g(H)$ with a simple count.

\section{Encoders and Thresholds}
\label{app:encoders}
Default encoder is \texttt{all-MiniLM-L6-v2}. We also test BGE and E5 families. Threshold $\delta$ is encoder‑specific; a sweep over $\delta\in[0.70,0.90]$ identifies a broad plateau where ACC is stable but RR@$\,\delta$ decreases with smaller $\delta$.
We recommend selecting the smallest $\delta$ that improves Uniqueness without harming ACC on a dev split.

\section{Output Schema and Preprocessing}
\label{data_process}
We adopt the XML-like schema in Eq.~\eqref{eq:structure} and enforce it at prompting and evaluation time. Concretely, the model is instructed to produce:
\begin{verbatim}
<strategy id="1">
  <reasoning> ... </reasoning>
  <strategy_outcome> ... </strategy_outcome>
</strategy>
...
<final_answer> ... </final_answer>
\end{verbatim}

\noindent\textbf{Validity.} A strategy block is \emph{valid} iff both
\texttt{<reasoning>} and \texttt{<strategy\_outcome>} are present and nonempty
(after trimming whitespace). We ignore malformed or duplicated blocks and keep
the remaining strategies in the order they appear, yielding $S(a)=\{(r_i,o_i)\}$ and
$n_{\text{strat}}(a)$ as in Sec.~\ref{sec:method}.

\noindent\textbf{Answer extraction.} The final answer is taken in the priority order
$\FA\ \succ \ANS\ \succ last \SO$ (Sec.~\ref{sec:method}). 
For all answer comparisons, we apply numeric canonicalization $N(\cdot)$.

\noindent\textbf{Preprocessing.} Before parsing, we normalize Unicode punctuation, collapse
repeated whitespace/newlines, and strip any spurious markup inside tags (e.g., Markdown fences).
Empty or ill-formed tags are dropped. The same parser is used during training (to compute
$R_{\mathrm{oc}},R_{\mathrm{re}},R_{\mathrm{fa}},R_{\mathrm{sd}}$) and during evaluation
(ACC, S-ACC, and diversity metrics), ensuring consistency between rewards and metrics.

\section{Hyperparameter Grids}
\label{app:hparams}
\begin{table}[hbt!]
\centering
\begin{tabular}{ll}
\toprule
\textbf{Parameter} & \textbf{Grid} \\
\midrule
$w_{\mathrm{oc}}, w_{\mathrm{re}}, w_{\mathrm{fa}}, w_{\mathrm{sd}}$ & $\{0.5, 1, 2\}$ each \\
$\beta$ (KL) & $\{0.01, 0.02, 0.05, 0.1\}$ \\
$G$ (group size) & $\{4, 6, 8\}$ \\
$\delta$ & $\{0.70, 0.75, 0.80, 0.85, 0.90\}$ \\
$\gamma_{\mathrm{correct}}$ & $\{0.5, 1.0, 1.5\}$ \\
$\gamma_{\mathrm{cap}}$ & $\{0.3, 0.5, 0.7\}$ \\
$\gamma_{\mathrm{rate}}$ & $\{0.05, 0.10, 0.20\}$ \\
LoRA $r$ & $\{32, 64\}$ \\
Max seq.\ length & $\{2048, 3072, 4096\}$ \\
\bottomrule
\end{tabular}
\caption{Hyperparameter search space.}
\end{table}

\begin{table}[hbt!]
\centering
\small
\setlength{\tabcolsep}{6pt}
\begin{tabular}{lc}
\toprule
\textbf{Model (GSM8K)} & \textbf{ACC (\%)} \\
\midrule
Base: \texttt{Qwen2.5-3B}        & 54.66 \\
Base + $R_{oc}$ + $R_{fa}$       & 75.15 \\
Base + $R_{oc}$ + $R_{fa}$ + $R_{sd}$       & 80.02 \\
\textbf{\method\ } (Base + $R_{oc}$ + $R_{fa}$ + $R_{sd}$ + $R_{re}$)  & \textbf{82.03} \\
\bottomrule
\end{tabular}
\caption{\textbf{GSM8K} Ablations with Qwen2.5-3B.}
\label{tab:gsm8k_acc_only}
\end{table}

\section{Ablations (GSM8K, Qwen2.5-3B)}
\label{subsec:ablations}

We conduct ablation studies on \textbf{GSM8K} using \texttt{Qwen2.5-3B} to isolate the contribution of each reward component introduced in Sec.~\ref{subsec:reward_comp}.  
Table~\ref{tab:gsm8k_acc_only} reports accuracy (ACC) under progressively enriched reward configurations.

Starting from the base model (54.66\% ACC), adding the outcome-consistency and format-adherence rewards ($R_{\mathrm{oc}} + R_{\mathrm{fa}}$) yields a substantial improvement to 75.15\%, demonstrating the importance of enforcing structural correctness and answer validity. Incorporating the semantic diversity reward ($R_{\mathrm{sd}}$) further improves accuracy to 80.02\%, indicating that \emph{semantic exploration} plays a critical role in discovering higher-quality reasoning trajectories.

Finally, introducing the remaining exploitation-related reward ($R_{\mathrm{re}}$) recovers the full \method\ (SD-E$^{2}$) objective, achieving the best performance at \textbf{82.03\%}. This final gain, while smaller in magnitude, confirms that controlled consolidation complements semantic exploration by stabilizing learning and preventing redundant reasoning patterns.

% Overall, the ablation results show that performance gains are driven primarily by improved exploration quality via $R_{\mathrm{sd}}$, while exploitation-aware rewards provide additional but consistent improvements when layered on top.

\end{document}